\documentclass{article}

\usepackage{PRIMEarxiv}

\usepackage[utf8]{inputenc} % allow utf-8 input
\usepackage[T1]{fontenc}    % use 8-bit T1 fonts
\usepackage{hyperref}       % hyperlinks
\usepackage{url}            % simple URL typesetting
\usepackage{booktabs}       % professional-quality tables
\usepackage{amsfonts}       % blackboard math symbols
\usepackage{nicefrac}       % compact symbols for 1/2, etc.
\usepackage{microtype}      % microtypography
\usepackage{lipsum}
\usepackage{fancyhdr}       % header
\usepackage{graphicx,subcaption}       % graphics
\graphicspath{{media/}}     % organize your images and other figures under media/ folder
\usepackage{amssymb}
\usepackage{amsmath}
\usepackage{multirow}
\usepackage{multicol}
\title{BYEL : Bootstrap Your Emotion Latent
}

\author{
  Hyungjun Lee, Hwangyu Lim \\
  Graduate School of Automotive Engineering, Kookmin University \\
  Seoul\\
  \texttt{\{rhtm13, yooer\}@kookmin.ac.kr} \\
   \And
  Sejoon Lim \\
  Department of Automobile and IT Convergence, Kookmin University \\
  Seoul\\
  \texttt{lim@kookim.ac.kr} \\
}

\begin{document}
\maketitle

\begin{abstract}
With the improved performance of deep learning, the number of studies trying to apply deep learning to human emotion analysis is increasing rapidly. But even with this trend going on, it is still difficult to obtain high-quality images and annotations. For this reason, the Learning from Synthetic Data (LSD) Challenge, which learns from synthetic images and infers from real images, is one of the most interesting areas. In general, Domain Adaptation methods are widely used to address LSD challenges, but there is a limitation that target domains (real images) are still needed. Focusing on these limitations, we propose a framework Bootstrap Your Emotion Latent (BYEL), which uses only synthetic images in training. BYEL is implemented by adding Emotion Classifiers and Emotion Vector Subtraction to the BYOL framework that performs well in Self-Supervised Representation Learning. We train our framework using synthetic images generated from the Aff-wild2 dataset and evaluate it using real images from the Aff-wild2 dataset. The result shows that our framework (0.3084) performs 2.8\% higher than the baseline (0.3) on the macro F1 score metric.
\end{abstract}

% keywords can be removed

\section{Introduction}
Human emotion analysis is one of the most important fields in human-computer interaction. With the development of deep learning and big data analysis, researches on human emotion analysis using these technologies are being actively conducted \cite{kollias2021distribution,kollias2021affect,kollias2020deep,kollias2020va,kollias2019expression,kollias2019deep,kollias2018photorealistic,kollias2017recognition}. In response to this trend, three previous Affective Behavior Analysis in-the-wild (ABAW) competitions were held in conjunction with the IEEE Conference on Face and Gesture Recognition (IEEE FG) 2021, the International Conference on Computer Vision (ICCV) 2021 and the IEEE International Conference on Computer Vision and Pattern Recognition (CVPR) 2022 \cite{kollias2020analysing,Kollias_2021_ICCV,kollias2022abaw}. The 4th Workshop and Competition on ABAW, held in conjunction with the European Conference on Computer Vision (ECCV) in 2022 comprises two challenges \cite{kollias2022abaw}. The first one is Multi-Task-Learning (MTL), which simultaneously predicts Valence-Arousal, Facial Expression, and Action Units. The second one is Learning from Synthetic Data (LSD), which trains with synthetic datasets and infers real datasets.

Due to the successful performance of deep learning, there have been many studies using it to perform human emotion analysis\cite{kollias2021distribution,oh2021causal,jeong2022multitask}. However, in human emotion analysis using deep learning, a large amount of high-quality facial datasets are required for successful analysis. The problem is that it is difficult to easily utilize such datasets in all studies because the cost of collecting a large number of high-quality images and their corresponding labels is high. Therefore, LSD Challenge, which utilizes synthetic datasets to train a neural network and to apply real datasets to the trained neural network, is one of the most interesting areas.

In this paper, we solve the LSD Challenge of ABAW-4th \cite{kollias2022abaw}. A prominent problem to be solved for the LSD Challenge is that the domain of training and inference is different. To solve this problem, Domain Adaptation (DA) techniques are commonly used. DA is a method that increases generalization performance by reducing the domain gap in the feature space of the source and target domains. Traditional DA methods reduce the domain gap in the feature space of the source domain and target domain using the adversarial network \cite{ganin2015unsupervised,tzeng2017adversarial,hoffman2018cycada}. Furthermore, studies have recently been conducted to reduce the gap between source and target domains in feature space using the characteristics of self-supervised learning (SSL) that learn similar representations in feature space without adversarial networks \cite{jain2022self,akada2022self}. However, both traditional DA and SSL-based DA have limitations in that both the source domain dataset and target domain dataset are necessary for the training phase. Focusing on these limitations, we propose an SSL-based novel framework that learns the emotional representation of the target domain(real images) using only the source domain(synthetic images).  Our contributions are as follows.

\begin{itemize}
    \item First, we propose the enabled emotion aware Self-Supervised Learning method to learn an invariant features that represent emotion in both the synthetic image and the real image. 
    \item Second, we solve domain adaptation by learning the optimal representation that is also applied in real images using only the synthetic image.
\end{itemize}
We confirm the efficiency by comparing our contributions with the methods of various cases in \ref{sec:results}.

\section{Related Work}
\subsection{Self-Supervised Representation Learning}
 \quad Recently, studies on methodologies for extracting representations using self-supervision are being actively conducted. MoCo \cite{he2020momentum} performs contrastive learning based on dictionary look-up. When the key and query representations are derived from the same data, learning is carried out in the direction of increasing the similarity. SimCLR \cite{chen2020simple} is proposed as an idea to enable learning without an architecture or memory bank. it learns representations to operate as a positive pair of two augmented image pairs.
 
All existing contrastive learning-based methodologies before BYOL use negative pairs. BYOL \cite{grill2020bootstrap} achieved excellent performance through a method that does not use negative pairs by using a method that utilizes two networks instead of using a negative pair. In this study, an online network  predicts the representation of target network which has same architecture with online network and updates the parameters of the target network using an exponential moving average. As such, the iteratively refining process is bootstrapping.

\subsection{Human Emotion Analysis}
\quad Human Emotion Analysis is rapidly growing as an important study in Human-computer interaction field. In particular, through the Affective Behavior Analysis in-the-wild(ABAW) competition, many methodologies are proposed and their performance has been improved. In the 3rd Workshop and Competition on ABAW, the four challenges i) uni-task  Valence-Arousal Estimation, ii) uni-task Expression Classification, iii) uni-task Action Unit Detection, and iv) Multi-Task Learning and evaluation are described with metrics and baseline systems \cite{kollias2022abaw}.

Many methodologies have been presented through the ABAW challenge. D. Kollias \textit{et al.} \cite{kollias2019deep,kollias2017recognition} exploits convolutional features while modeling the temporal dynamics arising from human behavior through recurrent layers of CNN-RNN from AffwildNet. They perform extensive experiments with CNNs and CNN-RNN architectures using visual and auditory modalities. and show that the network achieves state-of-the-art performance for emotion recognition tasks \cite{kollias2019expression}. According to one study \cite{kollias2021affect}, new multi-tasking and holistic frameworks are provided to learn collaboratively, generalize effectively. In this study, multi-task DNNs, being trained on AffWild2 outperform the state-of-the-art for affect recognition over all existing in-the-wild databases. D. Kollias \textit{et al.} \cite{kollias2021distribution} present FacebehaviorNet and perform zero- and few-shot learning to the ability to encapsulate all aspects of facial behavior.
MoCo \cite{he2020momentum} is also applied in the field of Human Emotion Analysis. EmoCo \cite{sun2021emotion}, an extension of the MoCo framework, removes non-emotional information in the features with the Emotion classfier, and then performs emotion-aware contrastive learning through intra-class normalization in an emotion-specific space.

 Also, various new approaches for facial emotion synthesis have been presented. D. Kollias \textit{et al.} \cite{kollias2018photorealistic} propose a novel approach to synthesizing facial effects based on 600,000 frame annotations from the 4DFAB database in terms of valence and arousal. VA-StarGAN \cite{kollias2020va} applies StarGAN to generate a continuous emotion synthesis image. D. Kollias \textit{et al.} \cite{kollias2020deep} propose a novel approach for synthesizing facial affect. In this study, impact synthesis is implemented by fitting a 3D Morphable Model to a neutral image, then transforming the reconstructed face, adding the input effect, and blending the new face and the given effect to the original image.

\begin{figure}[tb!]
  \centering
  \includegraphics[width=0.7\textwidth]{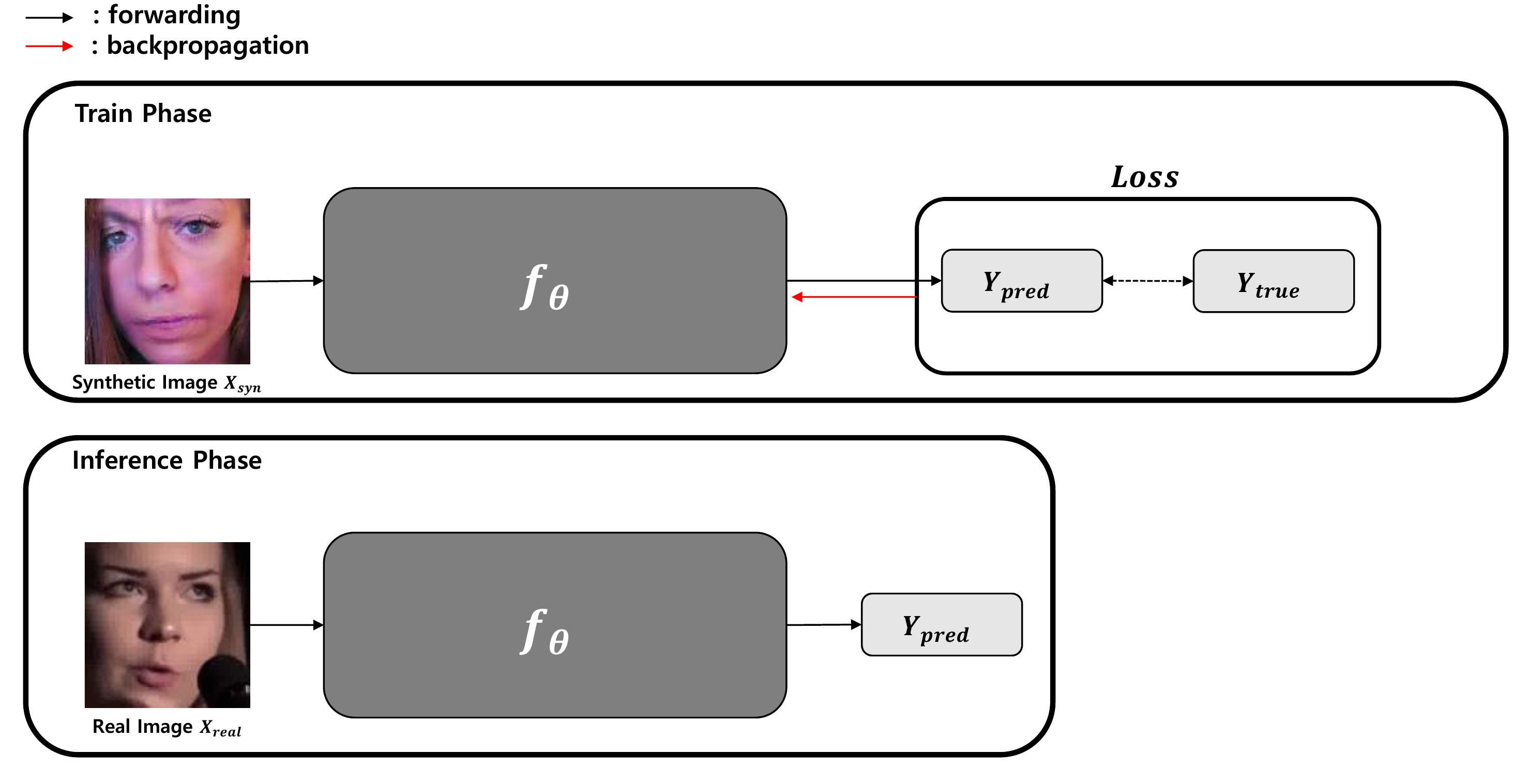}
  \caption{Problem description of ABAW-4th's LSD(Learning from Synthetic Data) Challenge \cite{kollias2022abaw}.}
  \label{fig:overview}
\end{figure}

\section{Problem Description}
\label{sec:problem_description}
\quad  ABAW-4th's Learning from Synthetic Data (LSD) Challenge is a task that uses synthetic datasets to train neural networks and classify emotions using trained neural networks in real images. In training phase, we train neural networks $f_{\theta}$ that classify emotions using $Y_{true}\in$ \{Anger, Disgust, Fear, Happiness, Sadness, Surprise \} corresponding to synthetic image $X_{syn}\in\mathbb{R}^{N\times N}$, where $N$ is size of image. Also predicted emotions from $X_{syn}$ are defined as $Y_{pred}\in$ \{Anger, Disgust, Fear, Happiness, Sadness, Surprise \}. In inference phase, $Y_{pred}$ is obtained using real image $X_{real}\in\mathbb{R}^{N\times N}$. Figure \ref{fig:overview} shows our problem description. 

\begin{figure}[t!]
  \centering
  \subfloat[Bootstrap Your Emotion Latent(Pre-training Phase)]{\includegraphics[width=0.8\textwidth]{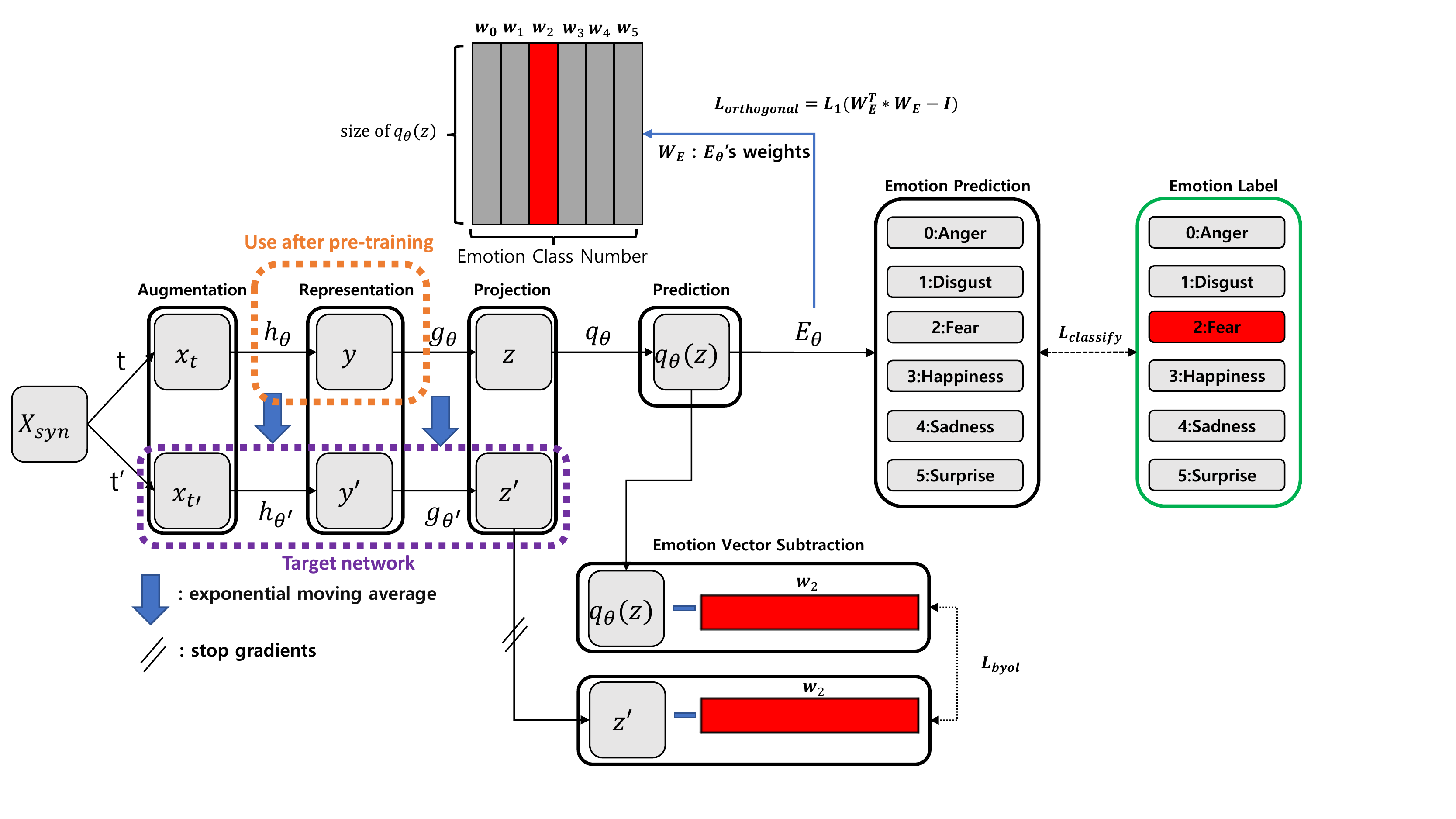}} \quad
  \subfloat[Transfer-learning Phase]{\includegraphics[width=0.6\textwidth]{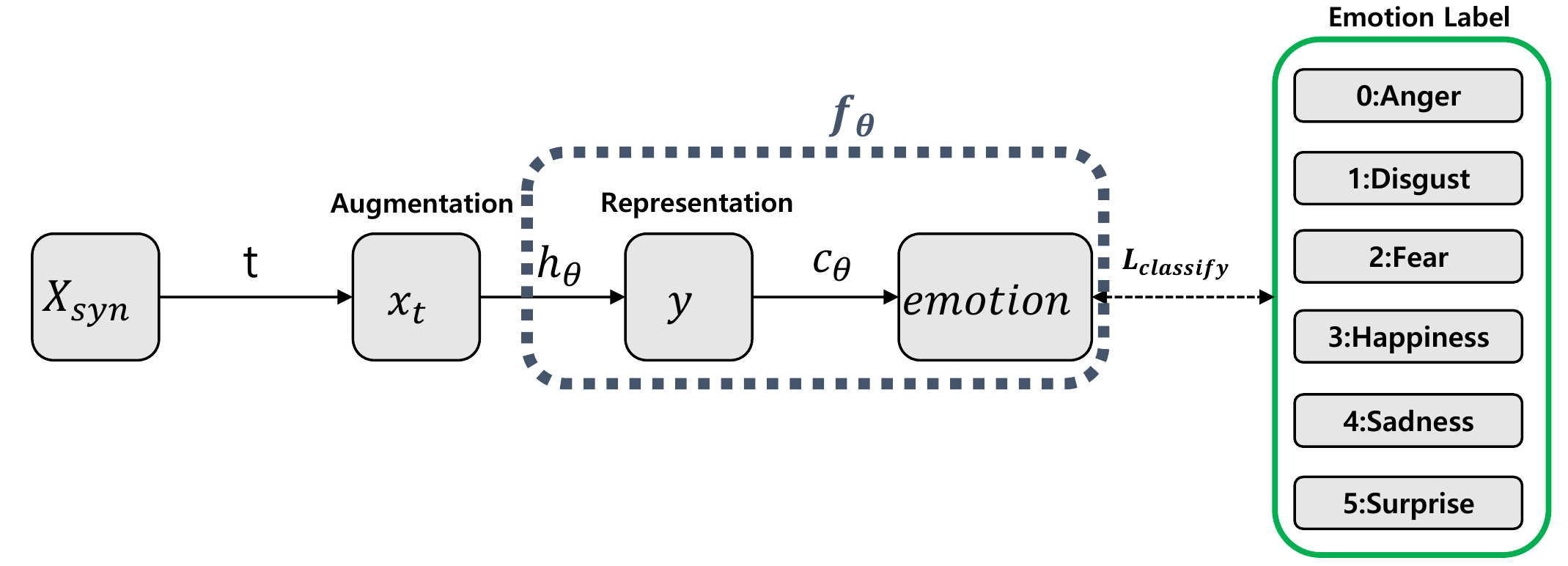}} 
  \caption{An illustration of our method.}
    \label{fig:model}
\end{figure}

\section{Method}
\quad Like previous Self-Supervised Learning frameworks, our method consists of two phases \cite{grill2020bootstrap,he2020momentum,chen2020simple}. The first, representation learning is conducted in the pre-training phase, and the second, transfer-learning is performed for emotion classification. We use the Bootstrap Your Emotion Latent (BYEL) framework to do representation learning and then transfer-learning for the emotion classification task. As shown in Figure \ref{fig:model} (a), the BYEL framework performs emotion-aware representation learning on feature extractor $h_{\theta}$. As shown in Figure \ref{fig:model} (b), $f_{\theta}$, which consists of pre-trained $h_{\theta}$ and classifier $c_{\theta}$, is trained in a supervised learning method in the emotion classification task. The final model, $f_{\theta}$, is formulated as equation \ref{equation_f}, where $\circ$ is the function composition operator.

\begin{center}
\begin{equation}\label{equation_f}
\begin{aligned}
    \mathit{f_{\theta}} = \mathit{c_{\theta}} \circ \mathit{h_{\theta}}(\circ:\textit{function composition operator})
\end{aligned}
\end{equation}
\end{center}

\subsection{Bootstrap Your Emotion Latent}
\label{sec:byel}
\quad Inspired by the excellent performance of EmoCo \cite{sun2021emotion} with MoCo \cite{he2020momentum} applied in the face behavior unit detection task, we apply BYOL \cite{grill2020bootstrap} to solve LSD tasks. There are several changes in applying BYOL to emotion-aware representaion learning. We add Emotion Classifier $E_{\theta}$ and Emotion Vector Subtraction.
\subsubsection{Emotion Classifier.} 
$E_{\theta}$ is a matrix with $W_{E}\in\mathbb{R}^{\textit{size of $q_{\theta}(z)$}\times C}$, where $C$ is number of emotion class. $W_{E}$ is a matrix that converts $q_{\theta}(z)$ into an emotion class. To conduct emotion-aware training as in EmoCo\cite{sun2021emotion}, we add Emotion Classifier $E_{\theta}$ to BYOL framework. The matrix $W_{E}$ is updated through the, $L_{classify}$, which is the Cross-Entropy of the Emotion Prediction and Emotion Label. As $W_{E}$ is trained, each column becomes a vector representing the corresponding emotion.
\subsubsection{Emotion Vector Subtraction.}  
Emotion Vector Subtraction is an operation to move  $q_{\theta}(z)$ to the emotion area within the feature space of  $q_{\theta}(z)$. Using $W_{E}$, we can obtain a prediction vector excluding the emotion information $\overline{q_{\theta}}(z)$ by subtracting the emotion vector $w_{idx}$ from the $q_{\theta}(z)$ of $X_{syn}$, like EmoCo \cite{sun2021emotion}. Here, $w_{idx}$ is a column vector of $W_{E}$ corresponding to the emotion label. In the same way, we subtract the emotion vector from the $z'$ of the target network to obtain the projection vector $\overline{z'}$ excluding the emotion vector $w_{idx}$. The whole process is formulated as equation \ref{emotion_vector}. 

\begin{center}
\begin{equation}\label{emotion_vector}
\begin{aligned}
    &\overline{q_{\theta}}(z)=q_{\theta}(z) - w_{idx} \\ &\overline{z'}=z' - w_{idx} \\ &(T:\textit{transpose}, idx \in \left\{0,...,C-1 \right\})
\end{aligned}
\end{equation}
\end{center}
 Figure \ref{fig:model} (a) shows the framework of BYEL, where feature extractor $h_{\theta}$, decay rate $\tau$, Projection layer $g_{\theta}$, Prediction layer $q_{\theta}$ and Augmentation function (t,t') is same as BYOL. The target network ($h_{\theta^{'}}$, $g_{\theta^{'}}$), which is label for representation learning, is not updated by $L_{byol}$ but only through exponential moving average of $h_{\theta}$, $g_{\theta}$ like BYOL. This target network update is formulated as an equation \ref{byel_update}. In addition, Figure \ref{fig:model} (a) is an example of a situation in which the emotion label is Fear. Here, since the label index of emotion vector corresponding to Fear in $W_{E}$ is 2, it can be confirmed that $w_{idx}$ is subtracted from  $q_{\theta}(z)$ and $z'$. After subtraction, as in BYOL, BYEL trains $\overline{q_{\theta}}(z)$ to have the same representation as $\overline{z'}$ so that $h_{\theta}$ performs emotion-aware representation learning for synthetic image $X_{syn}$.
\begin{center}
\begin{equation}\label{byel_update}
\begin{aligned}
    \mathit{\theta^{'}} = \tau*\mathit{\theta^{'}} + (1-\tau)*\mathit{\theta}(0\leq\tau\leq1)
\end{aligned}
\end{equation}
\end{center}

\subsection{Transfer-Learning}
\label{sec:transfer}
\quad After the pre-training phase, we can obtain $h_{\theta}$ with emotion-aware representation learning. Since $h_{\theta}$ can extracts emotion representation at $X_{syn}$, $f_{\theta}$ consists of feature extractor $h_{\theta}$ and classifier $c_{\theta}$, which is a one linear layer.  As shown in Figure \ref{fig:model} (b), $f_{\theta}$ is learned in the supervised learning method for the emotion classification task. 

\subsection{Loss}
\quad We use three loss functions to train our method. The first is $L_{classify}$ for emotion classification in pre-training phase, transfer-learning phase, the second is $L_{orthogonal}$ to orthogonalize the columns of $W_E$, and the third is $L_{byol}$ in pre-training phase. $L_{classify}$ is formulated as equation \ref{ce}, where $p$ is the softmax function and $y$ is the ground truth.
\begin{center}
\begin{equation}\label{ce}
\begin{aligned}
&\mathit{L_{classify}} = \mathit{-\sum_{c=0}^{C-1}y_c log(p(c))} (C:\textit{Class Number})
\end{aligned}
\end{equation}
\end{center}
Inspired by Pointnet's T-Net regularization \cite{qi2017pointnet}, which helps with stable training of transformation matrix, we use $L_{orthogonal}$ to train $E_{\theta}$ stably. $L_{orthogonal}$ is formulated as equation \ref{orthogonal}, where $I$ is identity matrix $\in \mathbb{R}^{C\times C}$ and $\left\| \cdot \right\|_1$ is the $L_1$ norm. 
\begin{center}
\begin{equation}\label{orthogonal}
\begin{aligned}
    &L_{orthogonal}=\sum_{i=0}^{C-1}\sum_{j=0}^{C-1}\left\|W_{E}^{T}*W_{E}-I \right\|_1[i][j]\\ 
    &(C:\textit{Class Number},I:\textit{Identity Matrix}\in \mathbb{R}^{C\times C})
\end{aligned}
\end{equation}
\end{center}
 $L_{byol}$ is the same as Mean Square Error with $L_2$ Normalization used by BYOL \cite{grill2020bootstrap}. $L_{byol}$ is formulated as equation \ref{byol}, where $\left< \cdot,\cdot \right>$ is the dot product function and $\left\| \cdot \right\|_2$ is the $L_2$ norm. 
 
\begin{center}
\begin{equation}\label{byol}
\begin{aligned}
    &L_{byol}=2-2\frac{\left< \overline{q_{\theta}}(z),\overline{z^{'}} \right>}{\left\| \overline{q_{\theta}}(z) \right\|_2 * \left\| \overline{z^{'}} \right\|_2}
\end{aligned}
\end{equation}
\end{center}
$L_{byel}$ is obtained by adding $\widetilde{L}_{byol}$, $\widetilde{L}_{classify}$ obtained by inverting t and t' in Figure \ref{fig:model} (a) to $L_{byol}$, $L_{classify}$, $L_{orthogonal}$ as in BYOL. Finally, $L_{byel}$ used in pre-training phase is formulated as equation \ref{total_loss} and Loss used in transfer-learning phase is formulated as equation \ref{ce}.
\begin{center}
\begin{equation}\label{total_loss}
\begin{aligned}
    L_{byel}=L_{byol}+\widetilde{L}_{byol}+L_{classify}+\widetilde{L}_{classify}+L_{orthogonal}
\end{aligned}
\end{equation}
\end{center}

\section{Experiments}
\label{sec:experiments}
\subsection{Dataset}
\quad Like the LSD task dataset in ABAW-4th \cite{kollias2022abaw}, synthetic images used in method development are all generated from real images used in validation. We can finally get a total of 277,251 synthetic images for training and a total of 4,670 real images for validation. Table \ref{tab:dataset} shows the detailed distribution of synthetic images and real images. Expression values are $\left\{0,1,2,3,4,5 \right\}$ that correspond to  \{Anger, Disgust, Fear, Happiness, Sadness, Surprise\}. 

\begin{table}[t!]
\centering
\caption{Distribution of datasets by emotion class.}
\begin{tabular}{@{}c|rr@{}}
\toprule
\multicolumn{1}{l|}{} & \multicolumn{2}{c}{\textbf{Number of Images}}                                              \\ \midrule
\textbf{Expression}   & \multicolumn{1}{c|}{\textbf{Synthetic Image}} & \multicolumn{1}{c}{\textbf{Real Image}} \\ \midrule
\textbf{0:Anger}      & \multicolumn{1}{r|}{18,286}                   & 804                                     \\ \midrule
\textbf{1:Disgust}    & \multicolumn{1}{r|}{15,150}                   & 252                                     \\ \midrule
\textbf{2:Fear}       & \multicolumn{1}{r|}{10,923}                   & 523                                     \\ \midrule
\textbf{3:Happiness}  & \multicolumn{1}{r|}{73,285}                   & 1,714                                   \\ \midrule
\textbf{4:Sadness}    & \multicolumn{1}{r|}{144,631}                  & 774                                     \\ \midrule
\textbf{5:Surprise}   & \multicolumn{1}{r|}{14,976}                   & 603                                     \\ \bottomrule
\end{tabular}
\label{tab:dataset}
\end{table}

\subsection{Settings}
\quad In the pre-training phase, we apply LARS \cite{you2017large} optimizer as in BYOL \cite{grill2020bootstrap} to train the BYEL framework and the $\tau$, augmentation t, projection layer $g_{\theta}$ and prediction layer $q_{\theta}$ are the same as BYOL \cite{grill2020bootstrap}, where epoch is 100, learning rate is 0.2, batch size is 256 and weights decay is $1.5-e^{-6}$. In transfer-learning phase, we apply Adam\cite{kingma2014adam} optimizer to learn the model $f_{\theta}$ consisting of $h_{\theta}$ that completed the 100-th epoch learning and 1 linear layer $c_{\theta}$, where epoch is 100, learning rate is $0.1-e^{-3}$ and batch size is 256. The size of images $X_{real},X_{syn}\in\mathbb{R}^{N\times N}$ is all set to $N=128$. We select a model with the best F1 score across all 6 categories(i.e., macro F1 score) after full learning. All experimental environments are implemented in pytorch \cite{NEURIPS2019_9015} 1.9.0.

\subsection{Metric}
\quad We use the evaluation metric F1 score across all 6 categories(i.e., macro F1 score) according to the LSD task evaluation metric proposed in ABAW-4th \cite{kollias2022abaw}. F1 score is defined as the harmonic mean of recall and precision and is formulated as equation \ref{metric}. Finally, the F1 score across all 6 categories (i.e., macro F1 score) is formulated as equation \ref{macro}. The closer the macro F1 score is to 1, the better the performance.

\begin{center}
\begin{equation}\label{metric}
\begin{aligned}
    Precision&=\frac{TruePositive}{TruePositive+FalsePositive} \\
    Recall&=\frac{TruePositive}{TruePositive+FalseNegative} \\
    F1-Score&=2*\frac{Precision*Recall}{Precision+Recall}
\end{aligned}
\end{equation}
\end{center}

\begin{center}
\begin{equation}\label{macro}
\begin{aligned}
    P_{LSD}=\frac{\sum_{c=0}^{5}F_1^{c}}{6}
\end{aligned}
\end{equation}
\end{center}

\subsection{Results}
\label{sec:results}
\quad We demonstrate the effectiveness of our method through comparison with the baseline presented in ABAW-4th \cite{kollias2022abaw}.  A baseline model is set to a transfer-learning model of ResNet50 \cite{he2016deep} pre-trained with ImageNet. ResNet50 with LSD is a case where ResNet50 is trained using the LSD dataset. BYOL with LSD is a case of training in the LSD dataset using the BYOL \cite{grill2020bootstrap} framework and then transfer-learning. BYEL with LSD is our method. Table \ref{tab:result} summarizes results. We also prove that our method is more effective than other methods.

\begin{table}[t!]
\centering
\caption{Comparison of macro F1 scores according to methods}
\begin{tabular}{@{}c|cc@{}}
\toprule
\textbf{}                  & \multicolumn{2}{c}{\textbf{macro F1 score with unit 0.01($\mathbf{\uparrow}$)}}                      \\ \midrule
\textbf{Method}            & \multicolumn{1}{c|}{\textbf{Validation set}} & \textbf{Test set} \\ \midrule
\textbf{baseline}          & \multicolumn{1}{c|}{50.0}                    & 30.0              \\
\textbf{ResNet50 with LSD} & \multicolumn{1}{c|}{59.7}                    & -                 \\
\textbf{BYOL with LSD}     & \multicolumn{1}{c|}{59.7}                    & 29.76             \\ \midrule
\textbf{BYEL with LSD}     & \multicolumn{1}{c|}{\textbf{62.7}}           & \textbf{30.84}    \\ \bottomrule
\end{tabular}
\label{tab:result}
\end{table}

\subsubsection{Ablation Study.} We analyze the relationship between pre-training epoch and performance through the performance comparison of $f_{\theta}^{e}=c_{\theta} \circ h_{\theta}^{e}$ according to the training epoch of pre-training. $h_{\theta}^{e}$ represents the situation in which training has been completed using the BYEL framework for as many as $e$ epochs. $f_{\theta}^{e}$ is a transfer-learned model using $h_{\theta}^{e}$. In Table \ref{tab:ablation}, it can be confirmed that the larger the pre-training epoch, the higher the performance. 

\begin{table}[t!]
\centering
\caption{Comparison of macro F1 scores in ablation study}
\begin{tabular}{@{}c|cc@{}}
\toprule
\textbf{}         & \multicolumn{2}{c}{\textbf{macro F1 score with unit 0.01($\mathbf{\uparrow}$)}}                      \\ \midrule
\textbf{Method}   & \multicolumn{1}{c|}{\textbf{Validation set}} & \textbf{Test set} \\ \midrule
\textbf{baseline} & \multicolumn{1}{c|}{50.0}                    & 30.0              \\
\textbf{$f_{\theta}^{45}$}       & \multicolumn{1}{c|}{56.9}                    & -                 \\
\textbf{$f_{\theta}^{90}$}       & \multicolumn{1}{c|}{59.3}                    & -                 \\ \midrule
\textbf{$f_{\theta}^{100}$}       & \multicolumn{1}{c|}{\textbf{62.7}}           & \textbf{30.84}    \\ \bottomrule
\end{tabular}
\label{tab:ablation}
\end{table}

\section{Conclusion}
\quad In this paper, inspired by EmoCo, we propose an emotion-aware representaion learning framework applying BYOL. This framework shows generalization performance in real images using only synthetic images for training. In section \ref{sec:results}, we demonstrate the effectiveness of our method. However, it does not show a very large performance difference compared to other methods. Therefore, we recognize these limitations, and in future research, we will apply the Test-Time Adaptation method to further advance.

%Bibliography
\bibliographystyle{unsrt}  
\bibliography{references}

\end{document}